\documentclass[sigconf,nonacm]{acmart}
\setkeys{acmart.cls}{balance=false}
\AtBeginDocument{%
  }

\usepackage{xspace}
\usepackage{subcaption}
\usepackage{stfloats}
\usepackage{booktabs}
\usepackage{array}
\usepackage{amsmath}

\newcommand{\model}{TaQFormer\xspace}

\begin{document}

\title[\model: Multidimensional Treatment Modeling for Coupon Allocation]{\model: A Treatment-as-Query Transformer with Data-Fusion for Multidimensional Treatment Modeling in Coupon Allocation}

\author{Jiahong Li}
\affiliation{%
  \institution{Didi Chuxing}
  \city{Beijing}
  \country{China}}
\email{caleblijiahong@didiglobal.com}

\author{Zeqin Yang}
\affiliation{%
  \institution{Didi Chuxing}
  \city{Beijing}
  \country{China}}
\email{yangzeqin@didiglobal.com}

\author{Yu Xiang}
\affiliation{%
  \institution{Didi Chuxing}
  \city{Beijing}
  \country{China}}
\email{yuxiangxiang@didiglobal.com}

\author{Zhifan Jiang}
\affiliation{%
  \institution{Didi Chuxing}
  \city{Beijing}
  \country{China}}
\email{zhifanjiang@didiglobal.com}

\author{Jingqi Ma}
\affiliation{%
  \institution{Didi Chuxing}
  \city{Beijing}
  \country{China}}
\email{irismajingqi@didiglobal.com}

\author{Enzheng Hua}
\affiliation{%
  \institution{Didi Chuxing}
  \city{Beijing}
  \country{China}}
\email{enzohua@didiglobal.com}

\author{Jixing Xu}
\affiliation{%
  \institution{Didi Chuxing}
  \city{Beijing}
  \country{China}}
\email{stevenxujixing@didiglobal.com}

\author{Jun Wu}
\affiliation{%
  \institution{Didi Chuxing}
  \city{Beijing}
  \country{China}}
\email{jackywujun@didiglobal.com}

\author{Fei Wang}
\affiliation{%
  \institution{Didi Chuxing}
  \city{Beijing}
  \country{China}}
\email{unsungwangfei@didiglobal.com}

\author{Enyun Yu}
\affiliation{%
  \institution{Didi Chuxing}
  \city{Beijing}
  \country{China}}
\email{franklinyu@didiglobal.com}

\author{Jiecheng Guo}
\affiliation{%
  \institution{Didi Chuxing}
  \city{Beijing}
  \country{China}}
\email{jasonguo@didiglobal.com}

\renewcommand{\shortauthors}{Li et al.}

\begin{abstract}
We study coupon allocation on an online marketplace offering multiple service categories. Issuing coupons requires choosing a vector of category-specific discounts, accounting for both their direct effects and cross-category demand shifts.
We present \model, a treatment-as-query Transformer for this multidimensional treatment setting.
Each treatment coordinate is represented as a query token, allowing cross-attention to estimate the response to the full treatment vector.
To combine large but policy-biased observational logs with a smaller stream of randomized controlled trial data, we use a two-stage pipeline: observational pretraining learns general outcome patterns, followed by LoRA fine-tuning on RCT data with a request-specific fusion gate.
We evaluate the resulting system through the checks required for production, including offline uplift metrics, treatment-parallel serving, latency monitoring, and a 14-day online A/B test.
At essentially the same subsidy cost rate, \model increases business value per user by 1.26\% over the production baseline and significantly outperforms both the production baseline and the RCT-only model, while meeting the latency budget. It was subsequently rolled out to all marketplace traffic and has operated at full scale for several months as of October 2026.
The deployment illustrates how multidimensional treatment modeling and observational pretraining, combined with adaptation using RCT data, can support effective joint discount decisions in production.
\end{abstract}


\maketitle

\section{Introduction}
We study discount coupon allocation on an online marketplace offering multiple service categories. A coupon's effect on the focal category can depend on the other discounts offered alongside it, as well as current prices, service quality, supply, and market conditions. For each request, the marketplace must estimate how alternative discount combinations affect completion in the focal category and decide which combination to offer. Uplift modeling supports this decision by estimating the incremental effects of alternative treatments, allowing the allocator to improve the business outcome for the focal category under a target subsidy cost rate. We evaluate \emph{business value per user}, which reflects the aggregate business contribution of completed transactions in the focal category. For confidentiality, we withhold its business-specific calculation and absolute values, and report relative changes between experimental groups.
\label{sec:intro-gaps}
Deploying uplift models for this setting raises three challenges.

\emph{Modeling a focal response to structured treatments.}
Each treatment is an $M$-dimensional vector of category-level discounts, and the number of combinations grows with the number of categories and discount levels. Treatments that assign the same discount to a given category share part of their structure, even when the remaining discounts differ. A model should exploit this shared structure while estimating how the full treatment vector affects completion in the focal category. Our previous production model, built on PEPNet~\cite{chang2023pepnet}, fuses embeddings of the discount coordinates into one treatment vector before applying gated prediction towers. It does not retain separate coordinate-level queries over the request representation. Treatment-as-query attention provides a starting point~\cite{zhang2023exploring}; we use it to estimate the focal response under the joint treatment.

\emph{Combining observational and RCT data.}
Observational (OBS) data cover a broad range of requests, but historical policies determine which treatments are assigned, producing uneven coverage of request--action pairs. Randomized controlled trial (RCT) data are available at a smaller scale and provide treatment comparisons within the randomized support. The challenge is to retain useful representations learned from OBS data while adapting treatment-response estimates with the smaller RCT dataset~\cite{wu2022integrative,cheng2021adaptive}.

\emph{Validating Transformer-based uplift models in production.}
Transformers have achieved strong results in industrial recommendation systems~\cite{gui2023hiformer,zhang2025onetrans}, motivating their use in structured uplift modeling. However, lower predictive loss or greater model capacity alone does not establish better uplift ranking or allocation outcomes. The challenge is to determine whether a Transformer-based model improves uplift quality on held-out RCT data and increases business value at a comparable subsidy cost rate in controlled online experiments, while meeting production serving requirements.

We present \model, a treatment-as-query Transformer that combines compositional treatment encoding with two-stage OBS--RCT training, and validate its effectiveness in production. Our main contributions are threefold:

(1) \textbf{Compositional queries for joint discount responses.}
Each discount coordinate has its own treatment token, allowing treatments with the same discount for a given category to share embedding parameters. A treatment decoder uses all treatment tokens to query the request representation. A nonlinear readout then combines the decoded tokens to predict completion in the focal category under the full treatment vector. The resulting scalar response estimate for each candidate action feeds into the subsidy allocator.

(2) \textbf{Randomized adaptation with a frozen observational reference.}
We preserve a frozen OBS-pretrained branch and adapt a second branch using low-rank adaptation (LoRA) and new prediction heads. A gate conditioned on the request and candidate action combines the two branches' logits under RCT supervision.

(3) \textbf{Production validation and business impact.}
We validate \model as a Transformer-based uplift model through offline uplift evaluation, production load tests, and a 14-day online A/B test. At essentially the same subsidy cost rate, \model increases business value per user by $1.26\%$ over the production baseline and also significantly outperforms the RCT-only model (both $p<0.05$). The deployed configuration scores all 33 feasible joint treatments with a P99 latency of $21\,\mathrm{ms}$ in a 500 QPS load test. The system was subsequently rolled out to all marketplace traffic.

\section{Related Work}

\paragraph{Uplift modeling with structured treatments.}
Uplift modeling estimates how an intervention changes an individual's outcome relative to a control condition. Early methods mainly use randomized control groups and focus on a single binary treatment~\cite{radcliffe2007using,gutierrez2017causal}. Later studies consider multiple treatments, more general outcomes, and treatment-specific costs~\cite{olaya2020survey,zhao2017multiple,zhao2019cost}. Neural HTE methods use treatment-specific outcome heads, smooth dose--response functions, or orthogonal objectives~\cite{shi2019adapting,lu2021vcnet,chernozhukov2022riesznet}. TransTEE supports discrete, continuous, and structured treatments, including multiple treatments acting simultaneously. It models treatment interactions with self-attention and uses treatment representations as queries over covariates~\cite{zhang2023exploring}. EFIN encodes treatment features, uses treatment-conditioned attention, and adds estimated uplift to a baseline response~\cite{liu2023explicit}. \model combines compositional discount tokens with a nonlinear readout of the decoded treatment representations to estimate the focal response. It also combines observational pretraining with randomized adaptation while preserving the same request format at serving time.

\paragraph{Combining randomized and observational data.}
Treatment assignment by historical policies can confound unadjusted outcome comparisons. Conditional estimation requires sufficient adjustment and overlap. RCT data provide randomized comparisons within the trial population and action support. Existing methods combine these sources through joint estimation, adaptive weighting, or representation transfer, addressing data efficiency and bias under their respective assumptions~\cite{wu2022integrative,cheng2021adaptive,hatt2022combining,triantafillou2023learning,degtiar2023combining}. Our focus is representation transfer and randomized adaptation for structured discount decisions: OBS pretraining supplies predictive information, while RCT supervision adapts predictions under randomized assignment, reducing reliance on the OBS unconfoundedness assumption stated in Section~\ref{sec:problem}.

\paragraph{Transformers, model capacity, and efficient industrial inference.}
Transformers have become a common tool for modeling interactions among tabular features. TabTransformer and FT-Transformer use self-attention to learn contextual feature representations~\cite{huang2020tabtransformer,gorishniy2021revisiting}. Industrial recommender systems apply related architectures to larger and more heterogeneous feature spaces. HiFormer models heterogeneous feature interactions, while Wukong studies how recommendation quality changes with model scale~\cite{gui2023hiformer,zhang2024wukong}. Recent systems further combine feature interaction with sequence modeling or replace full attention with more efficient token-mixing blocks~\cite{zhang2025onetrans,zhu2025rankmixer,jiang2026tokenmixer,ha2026unimixer,cai2026whale}. These studies show that model capacity and efficient inference are important design considerations for industrial prediction systems. However, most focus on ranking or response prediction rather than counterfactual estimation. Our capacity study instead examines how width and depth affect uplift quality under a fixed training budget, alongside the serving requirement to score all feasible treatment vectors.

\section{Problem Formulation}
\label{sec:problem}
\subsection{Coupled multi-treatment setting}

On the online marketplace, we allocate discount coupons for each request across $M$ substitutable service categories offered at the same time. The marketplace issues discount coupons whose category-specific discount depths form the selected treatment vector. Let $\mathcal M=\{1,\dots,M\}$ index these categories in decreasing order of base price. For each request $i$, $X_i=x_i$ is the context, which contains the features available online at decision time, and $A_i\in\mathcal A$ is the treatment. Each candidate treatment is an $M$-dimensional vector of discount depths:
\begin{equation}
    a=(a_1,\dots,a_M)\in\prod_{m\in\mathcal M}\mathcal D_m,
\end{equation}
where $a_m$ is the discount depth for category $m$, and a larger value means a deeper discount. All discount depths follow the same fixed-step grid:
\begin{equation}
    \mathcal D=\{d_k=d_{\min}+(k-1)\Delta:k=1,\dots,K\},
    \qquad \Delta>0.
\end{equation}
Each category has its own allowed range $\mathcal D_m\subseteq\mathcal D$, but all categories use the same step size. A business rule requires every higher-priced category to receive a strictly smaller discount than every lower-priced category. This rule defines the feasible action set:
\begin{equation}
    \mathcal A=
    \left\{a\in\prod_{m\in\mathcal M}\mathcal D_m:
    a_{m+1}-a_m\ge\Delta,\;m=1,\dots,M-1\right\}.
\end{equation}
Therefore, every feasible action satisfies $a_1<\cdots<a_M$. The reference action $a_0$ is the shallowest feasible action. Although this constraint removes many combinations from the full product grid, $|\mathcal A|$ still grows combinatorially with $M$ and the number of available discount depths. Throughout this paper, a \emph{treatment} refers to the full structured intervention with $M$ coordinates, while an \emph{action} refers to one feasible realization $a\in\mathcal A$.

\subsection{Causal target and identification}

We model exposure-to-completion conversion in a single focal service category. For each action
$a\in\mathcal A$, let $Y(a)\in\{0,1\}$ denote the potential completion
outcome for an eligible exposure: one if it leads to a completed transaction
in the focal category within the attribution window, and zero otherwise. All $M$ category-specific discounts remain part of the treatment
vector and can affect this response. Relative to a reference action $a_0$,
the conditional average treatment effect is
$\tau(x;a,a_0)=\mathbb E[Y(a)-Y(a_0)\mid X=x]$.
Let $\mu(x,a)=\mathbb E[Y(a)\mid X=x]$ denote the response surface.
Then $\tau(x;a,a_0)=\mu(x,a)-\mu(x,a_0)$, and
$\hat y(x,a)$ estimates this exposure-to-completion probability.

Identification from the OBS training data relies on the following causal assumptions. First, we assume
\emph{consistency}: if a request is assigned action $A=a$, its observed
outcome equals $Y(a)$. Second, we assume \emph{no interference}: one
request's outcome does not depend on the actions assigned to other requests.
Third, we assume \emph{conditional exchangeability} (unconfoundedness),
$\{Y(a):a\in\mathcal A\}\perp\!\!\!\perp A\mid X$. Finally, we assume
\emph{positivity}: for every covariate value $x$ in the OBS population and
every action $a\in\mathcal A$, $\Pr(A=a\mid X=x)>0$.

Under these assumptions, the response surface is identified as
$\mu(x,a)=\mathbb E[Y\mid X=x,A=a]$. Therefore,
$\tau(x;a,a_0)=\mathbb E[Y\mid X=x,A=a]
-\mathbb E[Y\mid X=x,A=a_0]$.
RCT fine-tuning uses randomized supervision within the trial support to adapt
response estimates and reduce reliance on the OBS unconfoundedness assumption.

\subsection{Decision formulation}

For request $i$ under action $a$, the allocator combines the predicted focal-category
exposure-to-completion probability $\hat y(x_i,a)$ with the expected business
contribution conditional on completion to estimate business value $\hat r_{ia}$.
The business-specific conversion-to-value rule is shared by all online
experimental groups and is omitted for confidentiality.

Let $c_{ia}$ denote the subsidy cost per unit of business value for request
$i$ under action $a$. The expected subsidy spend is then
$c_{ia}\hat r_{ia}$. Given a target subsidy cost rate $\rho$, the allocator
selects one action for each request by solving
\begin{align}
\label{eq:allocation-lp}
    \max_{b\in\mathcal B}\quad
    &\sum_{i\in\mathcal I}\sum_{a\in\mathcal A}
    b_{ia}\hat r_{ia} \\
    \text{s.t.}\quad
    &\sum_{i\in\mathcal I}\sum_{a\in\mathcal A}
    b_{ia}(c_{ia}-\rho)\hat r_{ia}\leq 0,
\end{align}
where $\mathcal I=\{1,\ldots,n\}$ and $\mathcal B$ is the set of binary
assignment matrices satisfying $\sum_{a\in\mathcal A}b_{ia}=1$ for every
request $i$. The constraint limits the aggregate expected subsidy cost rate
to at most $\rho$.

For a fixed nonnegative multiplier $\lambda$, the Lagrangian relaxation
separates across requests. We use the resulting score in the online decision
rule
\begin{equation}
\label{eq:dual-rule}
    a_i^\star
    =
    \arg\max_{a\in\mathcal A}
    \hat r_{ia}\bigl(1-\lambda(c_{ia}-\rho)\bigr).
\end{equation}
Offline, $\lambda$ is calibrated on logged traffic so that the realized
subsidy cost rate is close to $\rho$. Online, the model scores every action
in $\mathcal A$, and Eq.~\eqref{eq:dual-rule} selects the final treatment.
Thus, the action-response vector
$\{\hat y(x_i,a):a\in\mathcal A\}$ serves as the interface between the
model and the downstream allocator.

\section{The \model Model}
\label{sec:model}

\begin{figure*}[t]
  \centering
  \includegraphics[width=0.9\textwidth]{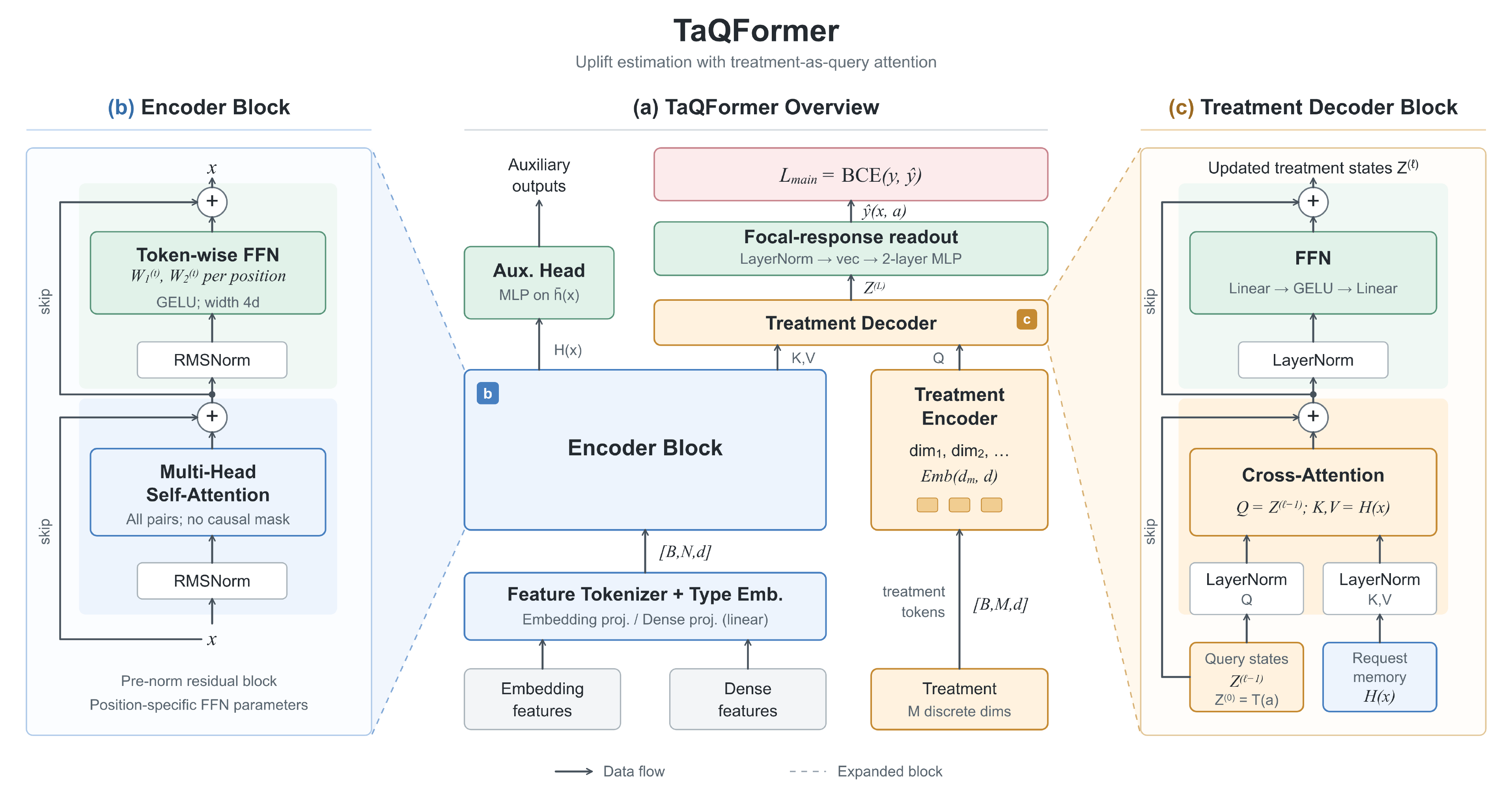}
  \caption{\model architecture: (a) focal-category response prediction with a treatment decoder followed by a separate focal-response readout, and an auxiliary branch; (b) pre-norm self-attention with token-wise FFNs; (c) a Treatment Decoder Block in which decoded treatment states query request memory.}
  \Description{Three-panel diagram of the \model architecture: the overall data flow from input features through the Encoder Block stack to the treatment decoder, focal-response readout, and auxiliary head; the internal structure of an encoder block; and the internal structure of one treatment decoder block.}
  \label{fig:architecture}
\end{figure*}

\subsection{Architecture}
\label{sec:model-arch}
\model uses two paths for response estimation: a treatment-independent request path and a treatment-dependent treatment path (Figure~\ref{fig:architecture}). For a serving context $x$, a feature tokenizer converts the input into $N$ semantic tokens, and a stack of Encoder Blocks (the Transformer backbone) encodes them into a request representation
$H(x)\in\mathbb R^{N\times d}$. For a feasible treatment $a$, each discount dimension is represented by a treatment token, giving $T(a)\in\mathbb R^{M\times d}$. These treatment tokens act as queries and use cross-attention to retrieve the relevant information from $H(x)$. A nonlinear readout combines the decoded treatment representations to predict the scalar completion probability $\hat y(x,a)$ for the focal category.
This scalar output estimates the probability of the primary binary label, which records whether an eligible exposure leads to completion under the assigned action. At serving time, Section~\ref{sec:serving} computes $H(x)$ once and reuses it to score every feasible action $a\in\mathcal A$, producing an action-response vector for each request.

\paragraph{Feature tokenizer and encoder.}
The tokenizer maps the serving-time context of Section~\ref{sec:problem} into a token sequence $X(x)\in\mathbb R^{N\times d}$: each scalar feature, categorical embedding, and pretrained structured vector receives its own token, with a shared projection within each input type and an added type embedding. Token positions have a fixed feature ordering. The tokens pass through the Encoder Block stack, whose pre-RMSNorm transformer blocks each interleave multi-head self-attention with a feed-forward layer (Figure~\ref{fig:architecture}b). The single departure from the vanilla block is that the feed-forward layers are \emph{token-wise}: each position owns private FFN parameters. Every token here is a fixed semantic field---a named feature field, not an interchangeable element of a sequence---so a shared FFN would impose one transformation on heterogeneous semantics, whereas per-token parameters give each field its own channel-mixing map, as in~\cite{zhang2025onetrans,zhu2025rankmixer}. The output is the action-independent request memory $H(x)$, which supplies keys and values to the treatment decoder; a pooled summary $\bar h(x)$ feeds an auxiliary head.

\paragraph{Compositional treatment encoding.}
\model retains a separate token for each treatment coordinate through the attention decoder, before combining the decoded coordinates in the focal-response readout. This differs from the production baseline, which fuses coordinate embeddings into one vector before gated prediction. For coordinate $j$, an embedding map
$e_j:\mathcal D_j\to\mathbb R^d$ converts the discount value $a_j$ into a vector. The treatment encoder then forms
\begin{equation}
    T(a) = [\,e_1(a_1),\dots,e_M(a_M)\,] \in \mathbb R^{M\times d}.
\end{equation}
This means that the embedding $e_j(d)$ is shared by every feasible treatment whose $j$th coordinate has value $d$, allowing related treatments to share information. At the same time, each coordinate has its own embedding table, so the model still knows which service category each discount belongs to.

\paragraph{Treatment-aware cross-attention.}
The treatment decoder stacks $L$ pre-norm Treatment Decoder Blocks, each comprising a cross-attention sublayer followed by a feed-forward sublayer (Figure~\ref{fig:architecture}c). It starts from $Z^{(0)}=T(a)$. At block $\ell$, the decoded treatment states $Z^{(\ell-1)}$ supply the queries, while the request memory $H(x)$ remains the keys and values. The cross-attention (CA) sublayer first produces a residual update,
\begin{equation}
    \label{eq:cross-attn-update}
    \widetilde Z^{(\ell)}
    = Z^{(\ell-1)} + \mathrm{CA}^{(\ell)}\!\bigl(\mathrm{LayerNorm}(Z^{(\ell-1)}),\, \mathrm{LayerNorm}(H(x))\bigr).
\end{equation}
For head $h$, with $Q_h^{(\ell)}$ and $K_h^{(\ell)}$ the projected normalized treatment and feature states, the coordinate-to-feature attention weights are
\begin{equation}
    P_h^{(\ell)}(x,a)=
    \mathrm{softmax}\!\left(
    \frac{Q_h^{(\ell)}\bigl(K_h^{(\ell)}\bigr)^\top}{\sqrt{d_h}}
    \right)\in\mathbb R^{M\times N}.
\end{equation}
Each row of $P_h^{(\ell)}$ corresponds to one treatment coordinate and retrieves the request features most relevant to that coordinate. Changing a discount can therefore change which market signals the corresponding treatment state attends to. The block then applies a second pre-norm residual update using its feed-forward sublayer,
\begin{equation}
    \label{eq:decoder-ffn-update}
    Z^{(\ell)}
    = \widetilde Z^{(\ell)} + \mathrm{FFN}^{(\ell)}\!\bigl(\mathrm{LayerNorm}(\widetilde Z^{(\ell)})\bigr).
\end{equation}
After $L$ blocks and a final LayerNorm, a separate two-layer GELU MLP jointly reads all $M$ decoded coordinates and produces the focal response,
\begin{equation}
\label{eq:readout}
    \begin{aligned}
    s(x,a) &= f\!\bigl(\mathrm{vec}(\mathrm{LayerNorm}(Z^{(L)}))\bigr), \\
    \hat y(x,a) &= \sigma\bigl(s(x,a)\bigr).
    \end{aligned}
\end{equation}
The joint readout $f$ combines the summaries from all $M$ treatment
coordinates. As a result, the focal response depends on both its own category
discount and the other discounts in the treatment vector. The decoder is also structurally aligned with the encoder. It operates on the full $N\times d$ request memory, uses the same hidden size and a pre-norm
residual structure (RMSNorm in the encoder and LayerNorm in the decoder), and does not first compress the request into a single
pooled vector. The Encoder Block therefore builds a treatment-independent representation of the request, while the Treatment Decoder Block adds treatment-specific interactions in an explicit and compositional way.

\subsection{Two-stage training: observational pretraining and randomized adaptation}
\label{sec:model-twostage}

\begin{figure*}[t]
  \centering
  \includegraphics[width=0.9\textwidth]{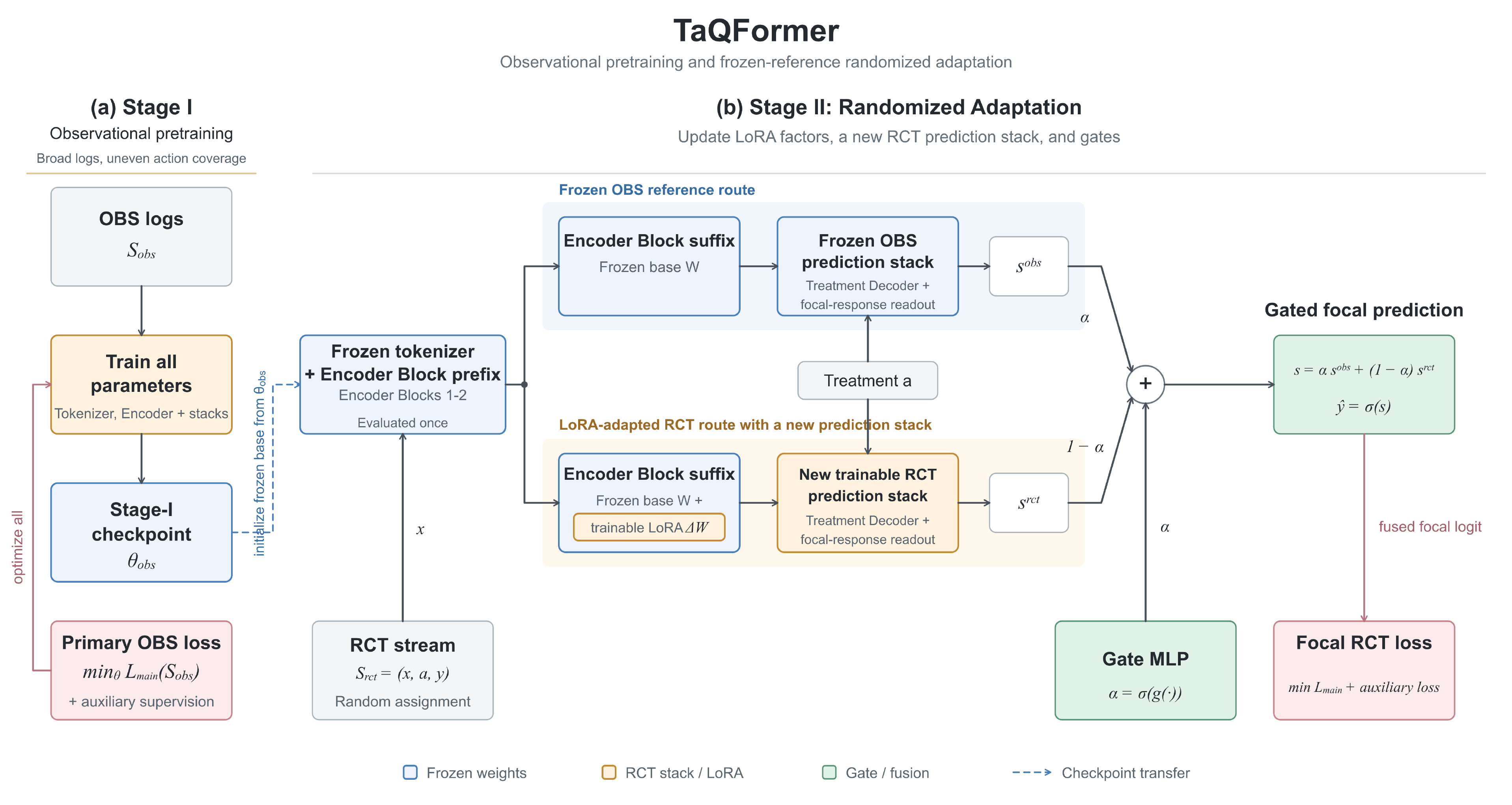}
  \caption{\model training: (a) OBS pretraining of all modules; (b) RCT adaptation with a frozen OBS reference, a shared frozen Encoder Block prefix, and a LoRA-adapted suffix. Only LoRA factors, a newly initialized RCT prediction stack (including its treatment encoder), and gates are updated. Gate inputs use stop-gradient; focal logits are fused, while the direct auxiliary output comes from the RCT stack.}
  \Description{Two-panel diagram of the two-stage training pipeline: observational pretraining of the full model, followed by randomized adaptation in which a frozen observational route and a newly initialized RCT prediction stack receive frozen-base and LoRA-adapted representations, respectively. An instance-wise gate fuses focal logits; the RCT stack provides the direct auxiliary output.}
  \label{fig:twostage}
\end{figure*}
Existing data-fusion methods learn from OBS and RCT data through representation transfer or prediction correction~\cite{hatt2022combining,demirel2024prediction,wu2022integrative,cheng2021adaptive}. With limited RCT data, our design aims to preserve a stable OBS prediction reference, limit the number of adapted parameters, and share computation at serving time. Throughout this section, for simplicity, a prediction stack comprises a treatment encoder, Treatment Decoder Blocks, a focal-response readout, and an auxiliary head. \model therefore retains a frozen OBS reference route while adapting an RCT route with LoRA and a newly initialized prediction stack (Figure~\ref{fig:twostage}). The comparison with full fine-tuning is evaluated empirically in Section~\ref{sec:twostage-eval}.

\paragraph{Stage I: observational pretraining.}
The full model---the tokenizer, Encoder Block stack, and prediction stacks---is trained for one epoch on OBS data using the primary binary cross-entropy objective of Eq.~\eqref{eq:primary-loss} together with auxiliary supervision. This stage learns the request representation and a treatment-aware decoder under the logging distribution. Uneven coverage, finite samples, and model misspecification can leave errors in the estimated response surface, motivating adaptation with randomized supervision.

\paragraph{Stage II: parameter-efficient randomized fine-tune.}
Stage II retains the complete OBS-trained model as an exact frozen reference route and constructs a distinct RCT route. The frozen tokenizer and first $L-L_{\mathrm a}$ Encoder Blocks form a shared prefix that is evaluated once. The final $L_{\mathrm a}$ Encoder Blocks are then evaluated twice: once with their original frozen weights to produce $H^{\mathrm{obs}}(x)$, and once with LoRA updates to produce $H^{\mathrm{rct}}(x)$. Trainable LoRA factors are attached to the attention projections and token-wise FFN factors in this suffix; all base Encoder Block weights remain frozen. This suffix-only design both limits the number of trainable parameters and avoids duplicating the early Encoder Block computation. For a frozen weight matrix $W$, the adapted route uses
\begin{equation}
   W' = W + (\eta / r) AB,
\end{equation}
where $A$ and $B$ are trainable rank-$r$ matrices and $\eta$ is a scaling constant. The frozen OBS prediction stack is copied from Stage I and maps $H^{\mathrm{obs}}(x)$ and frozen treatment tokens $T^{\mathrm{obs}}(a)$ to the reference focal logit $s^{\mathrm{obs}}(x,a)$ without computing gradients. In contrast, the newly initialized and trainable RCT prediction stack owns independent treatment tokens $T^{\mathrm{rct}}(a)$ and maps them with $H^{\mathrm{rct}}(x)$ to $s^{\mathrm{rct}}(x,a)$.

\paragraph{Instance-wise gated fusion.}
The paths' different training distributions and parameter constraints can yield different prediction errors across request--action pairs. Our gate parameterizes an instance-wise mixture of frozen and adapted focal logits, optimized for RCT outcome prediction. Its weights are not calibrated measures of causal reliability. A small MLP $g$ receives stop-gradient summaries of the adapted memory, the RCT treatment tokens, the two branch logits, and their absolute difference:
\begin{equation}
   \alpha(x,a) =
   \sigma\Bigl(
   g\bigl(
   \mathrm{sg}\bigl[
   \bar H^{\mathrm{rct}}(x),\,
   T^{\mathrm{rct}}(a),\,
   s^{\mathrm{obs}},\,
   s^{\mathrm{rct}},\,
   |s^{\mathrm{obs}}-s^{\mathrm{rct}}|
   \bigr]
   \bigr)
   \Bigr),
\end{equation}
where $\bar H^{\mathrm{rct}}(x)$ concatenates the mean- and max-pooled tokens of $H^{\mathrm{rct}}(x)$. The final logit is the gated average
\begin{equation}
   \label{eq:gated-fusion}
   \begin{aligned}
   s(x,a) &=
   \alpha(x,a)\,s^{\mathrm{obs}}(x,a)
   + \bigl(1-\alpha(x,a)\bigr)\,s^{\mathrm{rct}}(x,a),\\
   \hat y(x,a) &= \sigma\bigl(s(x,a)\bigr).
   \end{aligned}
\end{equation}

During Stage II, we optimize the same primary objective for one epoch on RCT data using the fused focal logit. We retain auxiliary supervision through the direct auxiliary output of the RCT prediction stack, which is not passed through the focal gate. The trainable parameters are therefore the LoRA factors, the RCT prediction stack, and the gates. The stop-gradient prevents the gate inputs from changing the representations indirectly, but gradients still reach $s^{\mathrm{rct}}$ through its explicit mixture weight in Eq.~\eqref{eq:gated-fusion}. The RCT route and gate remain jointly optimized through the fused prediction; stop-gradient blocks only the paths through the gate inputs.

\subsection{Treatment-parallel serving}
\label{sec:serving}

\model has an asymmetric train--serve interface. During training, each sample
contains only the action assigned by the historical or randomized policy, so
the model processes one $(x,a)$ pair. At serving time, however, the allocator
in Eq.~\eqref{eq:allocation-lp} needs predictions for all feasible actions,
$\{\hat y(x,a):a\in\mathcal A\}$, under a strict latency budget.

Our architecture separates the treatment-independent encoder from the
treatment-dependent decoder. For a request batch $X$ of size $B$, the encoder
runs once and produces $H(X)\in\mathbb{R}^{B\times N\times d}$. The feasible
treatment table is stored as a constant tensor and encoded as
$T(\mathcal A)\in\mathbb{R}^{|\mathcal A|\times M\times d}$. The treatment
dimension is introduced only in the decoder. We flatten the resulting
$B|\mathcal A|$ request--treatment pairs into one decoder batch and reshape the
outputs into a response matrix of size $B\times|\mathcal A|$. This
reuses the context representation across all actions and reduces the
per-request computation from
$|\mathcal A|(C_{\mathrm{enc}}+C_{\mathrm{dec}})$ to
$C_{\mathrm{enc}}+|\mathcal A|C_{\mathrm{dec}}$.

All actions are scored in one accelerator-friendly pass using static tensor
operations, without per-action loops or data-dependent control flow. The
two-stage model shares most of its computation as well. The tokenizer and the
first two Encoder Blocks form a common prefix, while only the final two
blocks are evaluated separately along the frozen OBS path and the
LoRA-adapted RCT path. The resulting memories,
$H^{\mathrm{obs}}(X)$ and $H^{\mathrm{rct}}(X)$, are each computed once and
then reused for all actions. Thus, the two-stage model duplicates only part
of the encoder rather than running the full encoder for every action.

\section{Experiments}
\label{sec:experiments}
We evaluate \model through offline experiments and online production tests,
organized around six research questions.
\textbf{RQ1: Overall effectiveness.}
Does \model outperform our production baseline and closely related
treatment-aware uplift models?
\textbf{RQ2: Architectural choices.}
Which components matter, and does the treatment-as-query decoder work with
alternative backbones?
\textbf{RQ3: Capacity selection.}
How do model width and depth affect predictive and uplift quality?
\textbf{RQ4: Two-stage training.}
Does OBS pre-training followed by RCT adaptation help, and do the fusion gate
and LoRA adaptation contribute?
\textbf{RQ5: Serving efficiency.}
How much does treatment-parallel inference reduce serving latency?
\textbf{RQ6: Online impact.}
How does \model perform in an online A/B test under production latency
constraints?

\subsection{Experimental Setup}
\label{sec:offline-eval}

We use offline experiments to answer RQ1--RQ4, a production load test to
answer RQ5, and an online A/B test to answer RQ6. For the architecture and capacity studies, all models are trained on the same RCT training set used for Stage II fine-tuning. Section~\ref{sec:twostage-eval} separately compares OBS-only, RCT-only,
mixed-data, and sequential training strategies, together with the
proposed two-stage pipeline. All single-stage runs use one training epoch, and each stage of the two-stage pipeline uses one epoch on its corresponding dataset. The industrial runs share the data split, seed, batch size, and GPU class (one NVIDIA RTX 6000D), and both industrial offline studies use the same held-out RCT set. Method-specific learning-rate schedules and auxiliary losses differ.

\paragraph{Metrics.}
For the industrial offline tables, we report Qini, AUC, and PCOC for the focal service category. For each of the
$|\mathcal A|-1=32$ non-reference actions, Qini ranks requests assigned to
that action and to the reference action $a_0$ by predicted uplift and computes
the normalized Qini coefficient. We report the average over all 32 actions.
AUC is the ROC AUC between the observed completion indicator and the
predicted completion probability under the assigned action. PCOC is the ratio of total predicted
probability to total observed outcomes; values close to one indicate good
aggregate calibration. Calibration is important because the downstream allocator uses
absolute response predictions to control subsidy spending.

\paragraph{Implementation details.}
The joint treatment spans $M=2$ service categories and
$|\mathcal A|=33$ feasible joint treatments. Each request is encoded once,
after which the focal-response decoder scores all 33 candidate treatments.
The encoder has dimension $d=64$, four Transformer blocks, four attention
heads, and feed-forward width $4d$. The treatment decoder has $L=2$ blocks.
In Figure~\ref{fig:scaling}, the experiment label \texttt{L2} refers to this
configuration: four Transformer blocks and two blocks per treatment decoder.
In Stage II, $L_{\mathrm a}=2$ adapts only the last two Transformer blocks; the
first two remain shared. The production LoRA rank
$r=16$ and scaling $\eta=16$. Stage I uses tens of millions of OBS records, while the RCT stream comprises millions of records.
These data scales reflect the deployed data regime rather than a
study of data-volume sensitivity.

The primary label $y$ is one if an eligible exposure leads to a completed
transaction in the focal category within the attribution window, and zero otherwise.
The model estimates the exposure-to-completion probability using a single
binary cross-entropy (BCE) objective:
\begin{equation}
\label{eq:primary-loss}
\mathcal L_{\mathrm{main}}(\mathcal S)
=
\sum_{(x,a,y)}
\ell_{\mathrm{BCE}}\!\left(y,\hat y(x,a)\right),
\end{equation} 
Auxiliary tasks are retained: their heads either use the decoder template
for request-level labels or read stop-gradient copies of the request and
treatment representations. Auxiliary supervision is used alongside the
primary objective, with missing auxiliary labels masked. Only the focal
completion prediction is passed to the allocator. We do not attribute the
reported gains to individual auxiliary tasks because no head-level ablation
is performed.

\subsection{RQ1: Overall effectiveness}

To answer RQ1, we compare \model with the production baseline, PEPNet
\cite{chang2023pepnet}, and two related treatment-aware models, EFIN
\cite{liu2023explicit} and TransTEE \cite{zhang2023exploring}. These industrial variants use the same features, factual labels, RCT data, and held-out metrics, and are trained directly on RCT data. Their complete training objectives differ: EFIN adds natural-response and intervention losses, while the production PEPNet retains propensity and pairwise auxiliary terms. PEPNet uses a shared MLP, fuses coordinate embeddings into one
treatment vector, and applies layer-wise gates to task-specific
towers. For EFIN, we represent treatment dimensions as tokens and replace its
single-layer additive attention with multi-head attention. For TransTEE, we
retain treatment-token cross-attention, add task-specific heads, and remove
propensity regularization because the data are randomized. \model achieves the best point estimates on all three metrics among PEPNet, EFIN, and TransTEE. Compared with
PEPNet, it raises Qini from 0.1612 to 0.2697 and AUC from 0.6776 to 0.7393,
while reducing PCOC from 1.3737 to 1.0012. EFIN improves over PEPNet but
remains substantially weaker, whereas TransTEE performs nearly as well as
\model but is slightly worse on every metric.

\begin{table}[t]
  \centering
  \small
  \setlength{\tabcolsep}{3pt}
  \caption{Single-stage comparison after one epoch on the randomized stream. Runs share data, seed, batch size, and GPU class and use the same held-out RCT week; learning-rate schedules and auxiliary losses differ. Higher is better for Qini and AUC; PCOC is best at 1.0. Bold marks the best value per column; alternative-encoder rows probe encoder choice and are excluded from bolding.}
  \label{tab:offline-main-results}
  \begin{tabular}{>{\raggedright\arraybackslash}p{1.25cm}>{\raggedright\arraybackslash}p{3.05cm}ccc}
    \toprule
    Group & Model & Qini & AUC & PCOC \\
    \midrule
    Baseline & PEPNet & 0.1612 & 0.6776 & 1.3737 \\
    \midrule
    External & EFIN (adapted) & 0.2129 & 0.6938 & 0.9802 \\
    External & TransTEE (adapted) & 0.2612 & 0.7362 & 1.0223 \\
    \midrule
    Ablation & flat MLP + Treatment Decoder & 0.2592 & 0.7184 & 1.1395 \\
    Ablation & Encoder Block + gated towers & 0.2536 & 0.7391 & 0.9446 \\
    \midrule
    Encoder & Wukong + Treatment Decoder & 0.2391 & 0.7051 & 0.9608 \\
    Encoder & UniMixer + Treatment Decoder & 0.2700 & 0.7385 & 0.9740 \\
    \midrule
    Ours & Encoder Block + Treatment Decoder & \textbf{0.2697} & \textbf{0.7393} & \textbf{1.0012} \\
    \bottomrule
  \end{tabular}
\end{table}

\subsection{RQ2: Architectural choices}

To answer RQ2, we conduct component ablations and encoder substitutions. Replacing
the Encoder Block with a flat MLP lowers Qini and AUC and worsens aggregate calibration
(Qini: 0.2697$\rightarrow$0.2592; AUC: 0.7393$\rightarrow$0.7184; PCOC:
1.0012$\rightarrow$1.1395). Replacing the Treatment Decoder Block with
PEPNet-style fused-treatment gated towers leaves AUC nearly unchanged but
reduces Qini to 0.2536 and changes PCOC to 0.9446. These bundled variants are consistent with a contribution from each block. However, the gated-tower variant also retains the production auxiliary losses, so its comparison does not isolate the decoder architecture alone. We also replace our encoder with Wukong and UniMixer while keeping the decoder
fixed. The alternative encoders use the same depth and hidden width, but their
parameter counts are not matched. The decoder remains effective with both,
showing that Treatment-as-Query is encoder-agnostic. Our encoder clearly
outperforms Wukong and performs comparably to UniMixer, with slightly better
AUC and aggregate calibration. Overall, the results support both the value of the
proposed blocks and the portability of the Treatment-as-Query decoder.

\begin{figure*}[t]
  \centering
  \includegraphics[width=0.85\textwidth]{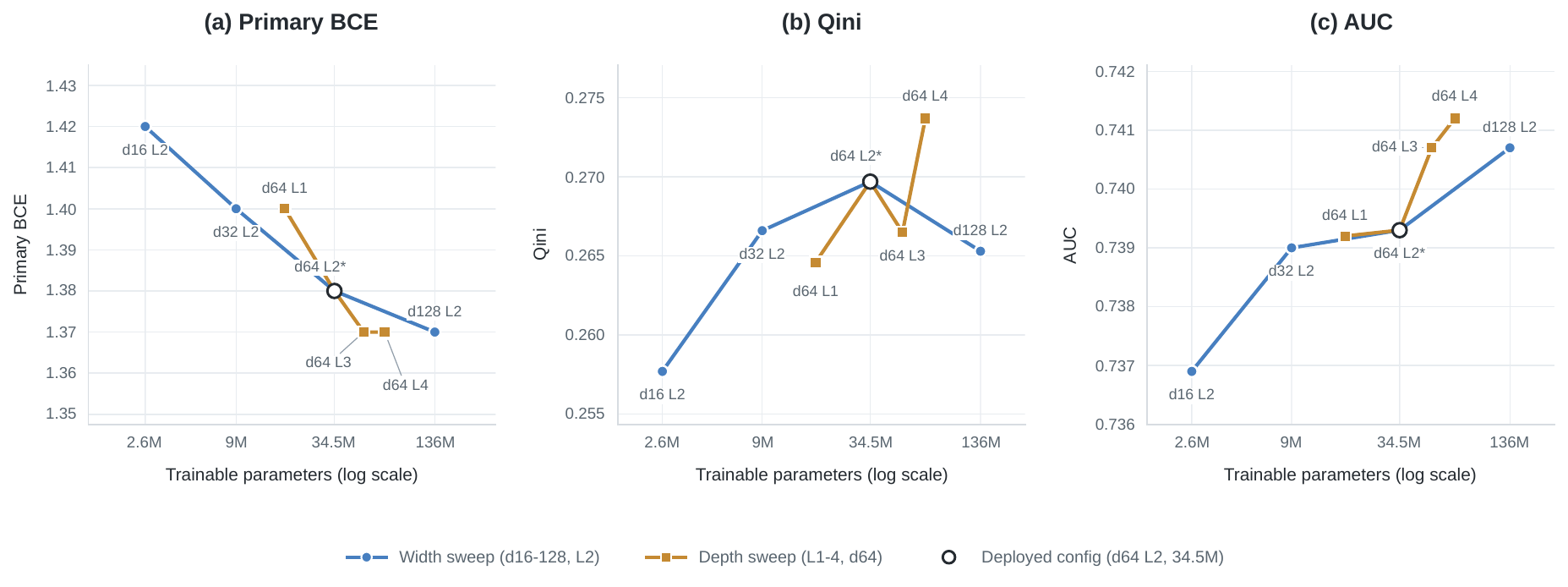}
  \caption{Single-stage \model capacity comparison after one RCT epoch. Width: $d_{\mathrm{model}}\in\{16,32,64,128\}$ at setting \texttt{L2}; depth settings: \texttt{L1}--\texttt{L4} use 2, 4, 6, and 8 Encoder Blocks and 1, 2, 3, and 4 Treatment Decoder Blocks, respectively, at $d_{\mathrm{model}}{=}64$. Open marker: deployed \texttt{L2} configuration (four backbone blocks, two blocks per treatment decoder; 34.5M parameters). The validation loss is the primary BCE. Single seed, one epoch.}
  \Description{Three panels compare primary BCE, Qini, and AUC across model widths and depth settings. Open markers identify the deployed configuration with four backbone blocks and two blocks per treatment decoder.}
  \label{fig:scaling}
\end{figure*}

\subsection{RQ3: Capacity selection}
\label{sec:scaling}

To answer RQ3, we compare seven models from 2.6M to 136M parameters,
varying width and depth around the deployed 34.5M-parameter configuration
(Figure~\ref{fig:scaling}). All configurations use one seed and one epoch
of RCT training. Validation loss is the primary BCE on the focal outcome.

Primary BCE generally decreased or plateaued as capacity increased, while
Qini did not consistently improve. Along the depth sweep, AUC rose from
0.7392 at setting \texttt{L1} to 0.7412 at \texttt{L4}, and Qini reached
0.2737 at \texttt{L4}, but dipped to 0.2665 at \texttt{L3}. Along the width sweep, widening beyond
$d_{\mathrm{model}}=64$ kept lowering primary BCE without improving
uplift ranking: the 136M-parameter width-128 model reached an AUC of 0.7407
but a Qini of only 0.2653, below the 0.2666 of the 9M-parameter width-32
model, at roughly the wall-clock cost of the 51.5M-parameter \texttt{L3} model.

These results inform capacity selection within the tested budget: lower
primary BCE alone is insufficient to select the model with the best uplift
ranking. We retain the deployed configuration as a practical capacity and
serving-cost choice.

\subsection{RQ4: Two-stage training}
\label{sec:twostage-eval}

To answer RQ4, Table~\ref{tab:twostage-ablation} compares training strategies
on the same held-out RCT set. The single-branch baselines retain the same
response-model architecture and use OBS-only training, RCT-only training,
jointly shuffled OBS and RCT data, or OBS pretraining followed by
full-parameter fine-tuning of all Stage-I modules on RCT data. The latter is a
conventional sequential baseline without a frozen reference branch or fusion
gate.

Shuffling OBS and RCT data improves all three metrics over OBS-only training,
and sequential full fine-tuning improves them further. However, the
sequential baseline still has lower Qini than RCT-only training
(0.2673 vs.\ 0.2697), despite its higher AUC (0.7421 vs.\ 0.7393).
Thus, adding OBS data does not automatically improve uplift ranking in
these runs. The proposed gated LoRA pipeline achieves the highest Qini
and AUC among these training strategies, improving both by 0.0033 over
conventional sequential full fine-tuning, while bringing PCOC closer to one (1.0114$\rightarrow$1.0027).
RCT-only training retains the PCOC closest to one (1.0012).

Within the frozen-reference variants, \emph{Encoder full fine-tuning + Gate}
replaces LoRA in the RCT Encoder Block suffix with full-parameter fine-tuning
of that suffix. \emph{LoRA + direct residual} keeps the LoRA-adapted RCT route
but replaces gated fusion with
$s(x,a) = s^{\mathrm{obs}}(x,a) + s^{\mathrm{rct}}(x,a)$.
Thus, gated fusion improves Qini and AUC over direct residual fusion with LoRA
fixed. With the gate fixed, LoRA has better point estimates than encoder full
fine-tuning on all three metrics while updating fewer parameters. These results
favor the selected configuration, but multi-seed uncertainty estimates are
unavailable. The comparison with conventional sequential fine-tuning evaluates
the complete adaptation design; it does not isolate an individual component.

\begin{table}[t]
  \centering
  \small
  \setlength{\tabcolsep}{4pt}
\caption{Training-strategy comparison on the same held-out RCT week as Table~\ref{tab:offline-main-results}. Single-branch rows share the response-model architecture; the lower group retains a frozen OBS reference. In the lower group, encoder full fine-tuning updates the RCT Encoder Block suffix directly, whereas direct residual uses $s=s^{\mathrm{obs}}+s^{\mathrm{rct}}$. Higher is better for Qini and AUC; PCOC is best at 1.0. Bold marks the best value in each column across all rows.}
  \label{tab:twostage-ablation}
  \begin{tabular}{@{}lccc@{}}
    \toprule
    Training variant & Qini & AUC & PCOC \\
    \midrule
    \multicolumn{4}{@{}l}{\emph{Single-branch training}} \\
    OBS only & 0.2548 & 0.7382 & 1.0315 \\
    RCT only & 0.2697 & 0.7393 & \textbf{1.0012} \\
    OBS + RCT (shuffled) & 0.2665 & 0.7398 & 1.0136 \\
    OBS $\rightarrow$ RCT (full fine-tuning) & 0.2673 & 0.7421 & 1.0114 \\
    \midrule
    \multicolumn{4}{@{}l}{\emph{Frozen-reference adaptation}} \\
Encoder full fine-tuning + Gate & 0.2700 & 0.7416 & 1.0100 \\
    LoRA + direct residual & 0.2686 & 0.7443 & 0.9923 \\
    LoRA + Gate (\model) & \textbf{0.2706} & \textbf{0.7454} & 1.0027 \\
    \bottomrule
  \end{tabular}
\end{table}

\subsection{RQ5: Serving efficiency}

To answer RQ5, we compare the proposed treatment-parallel implementation with
the original implementation, which reruns the full encoder for every action.
The serving configuration uses NVIDIA RTX 6000D hardware and a batch size of 256. We conduct a 10-minute load test at 500 QPS and report P99 serving latency.
For the two-stage model, treatment-parallel serving reduces P99 latency from
34\,ms to 21\,ms, a 38\% reduction. The same optimization reduces P99
latency for the one-stage RCT model from 29\,ms to 18\,ms. These results show
that reusing the encoded request context and evaluating all treatments in a
single static graph substantially reduces serving latency while preserving
the model used during training.

\subsection{RQ6: Online impact}
\label{sec:online-ab}

To answer RQ6, we ran a 14-day A/B test on live traffic. Each user was
placed in one group for the whole test, based on a hash of the user ID, so
the same user always saw the same model. The control group kept the
production PEPNet model; one treatment group used the single-stage model
trained only on RCT data, and the other used the two-stage OBS--RCT model.
Each group received $5\%$ of traffic, with sample sizes in the millions
for each group. All three groups shared the same
allocator and subsidy-rate target, and their realized cost rates were almost
identical: $20.43\%$, $20.45\%$, and $20.42\%$.

At essentially the same subsidy cost rate, business value per user rose by $1.08\%$ under the
single-stage model and by $1.26\%$ under the two-stage model, both relative
to the PEPNet control (Table~\ref{tab:online-ab}). The two-stage model
significantly outperforms both the production baseline and the RCT-only model.
All three pairwise comparisons are significant under CUPED-adjusted two-sample
$t$-tests ($p<0.05$). Relative to RCT-only, the two-stage model delivers
approximately $0.178\%$ additional business value per user, calculated from
the reported relative changes.

\begin{table}[t]
  \centering
  \small
  \setlength{\tabcolsep}{5pt}
  \caption{Online A/B results. Business value per user is indexed to PEPNet${}=100$. Superscripts $*$ and $\dagger$ indicate significant gains over PEPNet and RCT-only, respectively ($p<0.05$).}
  \label{tab:online-ab}
  \begin{tabular}{@{}lcc@{}}
    \toprule
    Model & Business value index & Subsidy rate \\
    \midrule
    PEPNet & 100.00 & 20.43\% \\
    RCT-only & $101.08^{*}$ & 20.45\% \\
    Two-stage (\model) & $101.26^{*\dagger}$ & 20.42\% \\
    \bottomrule
  \end{tabular}
\end{table}

The two-stage model was then rolled out to all traffic and has operated at
full scale for several months as of October 2026. This deployment duration
is separate from the 14-day A/B test. For confidentiality, the application
setting and the business-specific calculation are described in anonymized
terms, and absolute values are withheld; relative comparisons and
significance results are unchanged.

\section{Conclusion}
We presented \model, a treatment-as-query Transformer for joint discount coupon allocation across service categories on an online marketplace. It combines compositional treatment queries and a focal-response readout with two-stage OBS–RCT training that preserves a frozen observational reference. Offline evaluation shows improved uplift ranking and aggregate calibration for the focal category over the production baseline, while the complete two-stage pipeline further improves response prediction and online performance over RCT-only training. In a 14-day online A/B test, \model increased business value per user by 1.26\% over the production baseline at essentially the same subsidy cost rate and also significantly outperformed RCT-only training (both $p<0.05$). The deployed configuration scores all 33 feasible joint treatments with a P99 latency of 21 ms in a 500 QPS load test. The system was subsequently rolled out to all marketplace traffic. These results validate \model as an effective Transformer-based uplift model for joint discount coupon decisions in production.

We plan to extend joint coupon allocation to more than seven service categories. This expansion will require further evaluation of treatment coverage, candidate-action selection, and serving efficiency as the feasible action space grows.

\bibliographystyle{ACM-Reference-Format}
\bibliography{oneuplift_refs}

\end{document}